\documentclass[final,3p,authoryear]{elsarticle}
\usepackage[fleqn]{amsmath}

\usepackage{float,times}
\usepackage{amssymb,amsthm,mathrsfs}
\usepackage{lineno}
\usepackage{epsfig}
\usepackage{color}
\usepackage{setspace}
\usepackage{latexsym}
\biboptions{numbers,sort&compress}
\usepackage{algpseudocode}
\usepackage{wrapfig} 
\usepackage{flushend} 
\usepackage{subfig,graphicx,epstopdf}
\usepackage{exscale,relsize}
\usepackage{algorithm}
\usepackage[colorlinks=true,linkcolor=cyan,citecolor=cyan,urlcolor=cyan]{hyperref}
\usepackage[doipre={doi:~}]{uri}
\allowdisplaybreaks[4]

\newdefinition{lem}{Lemma }[section]
\newdefinition{dnt}{Definition}[section]
\newdefinition{rem}{Remark }[section]
\newdefinition{ass}{Assumption}[section]
\newdefinition{exam}{Example}[section]

\usepackage{booktabs}
\usepackage{makecell}
\makeatletter
\newenvironment{breakablealgorithm}
  {
   \begin{center}
     \refstepcounter{algorithm}
     \hrule height.8pt depth0pt \kern2pt
     \renewcommand{\caption}[2][\relax]{
       {\raggedright\textbf{\ALG@name~\thealgorithm} ##2\par}%
       \ifx\relax##1\relax 
         \addcontentsline{loa}{algorithm}{\protect\numberline{\thealgorithm}##2}%
       \else 
         \addcontentsline{loa}{algorithm}{\protect\numberline{\thealgorithm}##1}%
       \fi
       \kern2pt\hrule\kern2pt
     }
  }{
     \kern2pt\hrule\relax
   \end{center}
  }
\makeatother

\begin{document}

\begin{frontmatter}

\title{Resource-constrained Amazons chess decision framework integrating large language models and graph attention \tnoteref{t1}}

\tnotetext[t1]{This work was supported by the National Key Research and Development Project of China under Grant (No. 2020YFA0714300), National Natural Science Foundation of China under Grant (No. 61833005 and 12061088), the Open Project of Key Laboratory of Transport Industry of Comprehensive Transportation Theory (Nanjing Modern Multimodal Transportation Laboratory) under Grant MTF2023004, the China Postdoctoral Science Foundation under Grant 2024T170129 and Grant GZC20240261.}

\author[label1]{Tianhao Qian}\ead{qth2mir@seu.edu.cn}
\author[label1,label2]{Zhuoxuan Li}\ead{230229338@seu.edu.cn}
\author[label1]{Jinde Cao\corref{cor1}} \ead{jdcao@seu.edu.cn}
\author[label4]{Xinli Shi}\ead{xinli\_shi@seu.edu.cn}
\author[label2,label5,label6]{Leszek Rutkowski}\ead{leszek.rutkowski@ibspan.waw.pl}

\address[label1]{School of Mathematics, Southeast University, Nanjing, 210096, China}
\address[label2]{Systems Research Institute of the Polish Academy of Sciences, Warsaw 01-447, Poland}
\address[label4]{School of Cyber Science \& Engineering, Southeast University, Nanjing, 210096, China}

\address[label5]{ Institute of Computer Science, AGH University of Krakow, Kraków 30-059, Poland}
\address[label6]{SAN University, Łódź 90-113 , Poland}

\cortext[cor1]{Corresponding author}

\begin{abstract}
Intelligent game-playing systems have become important platforms for evaluating decision-making, strategic planning, and adaptive learning in artificial intelligence. Nevertheless, training competitive agents in resource-constrained environments remains difficult, mainly because conventional deep learning methods require large amounts of expert data and high computational cost. This paper presents a lightweight hybrid framework for the Game of the Amazons to investigate weak-to-strong generalization by integrating graph-based structural learning with large language model-based data generation. In the proposed framework, a Graph Attention Autoencoder is introduced to extract structural information and guide a depth-focused multi-step Monte Carlo Tree Search, while a Stochastic Graph Genetic Algorithm is developed to refine evaluation signals. Meanwhile, GPT-4o-mini is adopted to generate synthetic training data, enabling the framework to learn from noisy and imperfect supervision without relying on expert demonstrations. Experiments on a $10\times10$ Amazons board show that the proposed method achieves win rates of 57.5\%--79.5\% against ablation baselines under limited search budgets, along with 15\%--56\% improvements in decision accuracy. Moreover, the framework surpasses its teacher model, GPT-4o-mini, achieving a competitive win rate of  66.5\% with moderately increased search effort. The results indicate that graph-based structural filtering can effectively mitigate noise in LLM-generated supervision and support the development of specialized game agents from general-purpose foundation models under strict computational constraints.
\end{abstract}


\end{frontmatter}

\section{Introduction}\label{sec1}
Recent years have witnessed rapid advances in large language models (LLMs) and deep learning, substantially expanding the capabilities of intelligent decision-making systems \citep{zhao2023survey}. However, many high-performing AI paradigms still rely on substantial computational resources, large-scale data, and expensive training pipelines, which limits their practical deployment on everyday machines and in resource-constrained environments. As a result, there is growing interest in designing lightweight yet effective decision-making frameworks that can maintain competitive performance without requiring high-end hardware.

This trend has already appeared in several application domains. For example, Yang et al. \citep{yang2025risk} combined multimodal knowledge graphs with LLMs for risk decision-making, Song et al. \citep{song2025deep} investigated deep reinforcement learning for model deployment in edge environments, and Zhu et al. \citep{zhu2025task} studied task offloading using LLM-enhanced multi-agent reinforcement learning. In broader engineering informatics, the integration of generative autoencoders and graph attention networks has shown great promise in handling data scarcity and generating robust synthetic data representations \citep{vae_gat_2024}. Furthermore, hybrid frameworks combining LLMs with graph-based reasoning have been demonstrated to significantly improve decision accuracy by filtering out LLM hallucinations in complex structural management \citep{hybrid_llm_graph_2025}. Nevertheless, competitive game intelligence under limited computational budgets remains comparatively underexplored. If intelligent systems can achieve strong performance on ordinary devices, a broader community of users could participate in deployment and data generation, which would in turn accelerate the development of practical AI systems.

To study this problem, we consider the Game of the Amazons as a representative testbed. Amazons is a highly challenging board game with a large branching factor and long-horizon strategic dependencies, making it well suited for evaluating the effectiveness of intelligent search and decision-making algorithms \citep{song2014enhanced}. The move-and-block mechanism substantially enlarges the search space, and although several small board settings have been solved, the standard 10$\times$10 game remains unsolved within limited time budgets \citep{song2014enhanced}. These properties make Amazons an appealing benchmark for studying resource-constrained decision intelligence.

Despite being played on a finite 10$\times$10 board, Amazons induces an extremely large effective search space due to its two-stage action rule. In each turn, a player first moves one of four pieces along a straight line and then places a barrier following the same movement rule. The goal is to restrict the opponent until no legal move remains. As a result, decision-making in Amazons must balance immediate local advantage with long-term global control. Under resource constraints, this task becomes particularly challenging because the game suffers from severe combinatorial explosion: the number of legal moves in a standard position often reaches the hundreds or even thousands, which places heavy pressure on exhaustive or backward-pruning search strategies \citep{hearn2005amazons}. Although deeper search may improve move quality, its computational cost rises rapidly, making it impractical in lightweight settings.

Another major challenge lies in reliable state evaluation. In Amazons, a move that appears locally optimal for restricting the opponent may simultaneously expose the player’s own pieces to future danger, whereas a seemingly disadvantageous local action may prove beneficial to the overall outcome \citep{snatzke2004new}. Therefore, an effective agent must assess not only the current position but also the downstream consequences of candidate moves. However, pure search becomes increasingly expensive as look-ahead depth grows, while learning-based evaluation is hindered by the scarcity of high-quality historical game records. Compared with Go or Chess, Amazons has far fewer expert-level datasets, making it difficult to learn robust value functions through supervised or reinforcement learning alone. In addition, purely neural approaches often improve expressive power at the expense of interpretability, whereas handcrafted evaluation functions remain understandable but may fail to capture the complex structure of board states. These challenges indicate that an effective Amazons framework should jointly address search efficiency, evaluation reliability, structural modeling, and data scarcity, rather than relying on any single mechanism in isolation.

To address the above challenges, a lightweight hybrid decision-making framework for the Game of the Amazons is proposed. At the search level, conventional exhaustive trees are replaced with a Monte Carlo Tree Search (MCTS) structure to reduce computational burden. On this basis, a multi-round depth-focused search mechanism is introduced to suppress exponential tree growth and improve the effectiveness of deeper-node exploration under limited budgets.

At the evaluation and selection levels, two lightweight autoencoder layers are employed to refine evaluation parameters for movement and placement stages, respectively, with limited additional latency. Meanwhile, decision quality is improved from both stochastic and structural perspectives: the proposed Graph-Attention Autoencoder (GAT-AE) captures the graph structure induced by MCTS, while the Stochastic Graph Genetic Algorithm (SGGA) introduces controlled randomness into candidate sampling. To further reduce dependence on expert game logs, synthetic supervision is generated by a general-purpose LLM, and an update mechanism is designed to decouple objective values from their origins.

The main contributions of this paper are summarized as follows:
\begin{itemize}
    \item A lightweight hybrid framework for resource-constrained Amazons decision-making is proposed, in which MCTS, autoencoder-based evaluation refinement, SGGA-based stochastic candidate selection, and GAT-AE-based structural representation are integrated into a unified pipeline.

    \item A depth-focused multi-round search strategy is introduced to alleviate the conventional trade-off between search depth and computational cost, thereby enabling effective candidate exploration without exhaustive breadth expansion.

    \item A structural learning mechanism based on the graph topology induced by MCTS is developed. By combining stochastic graph selection with graph-attention-based filtering, candidate evaluation is improved beyond isolated node scoring.

    \item Weak-to-strong generalization without expert demonstrations is explored. GPT-4o-mini is employed as a weak supervisor to generate noisy synthetic signals, and it is shown experimentally that hallucinated or inconsistent supervision can be filtered while stronger task-specific strategies are learned.
\end{itemize}

\section{Related Work}\label{sec2}
Research on Amazons game AI has mainly followed two directions: search-based methods and learning-based methods. Owing to the large branching factor of the game, early studies primarily focused on combinatorial search and pruning strategies. Tong et al. \citep{tong2019research} developed a parallelised PVS framework equipped with a transposition table and historical-heuristic scoring, which improved search efficiency by pruning irrelevant branches more aggressively. Dehghani et al. \citep{dehghani2017ga} proposed a genetic-algorithm-based approach for reducing the effective search space, showing that its runtime grows much more gently than that of traditional Min-Max as search depth increases.

Besides these approaches, Fu et al. \citep{fu2018amazon} proposed a staged UCT algorithm in which Linear Discriminant Analysis \citep{fisher1936use} was used to compress board information and K-means clustering was adopted to divide different game stages, thereby enabling stage-dependent weighting strategies. These studies demonstrate the importance of search-space control in Amazons. However, many search-based approaches still rely heavily on handcrafted evaluation functions, which may not adequately characterize the subtle strategic structure of complex board states.

Learning-based methods have also been explored for Amazons. Jianbo et al. \citep{jianbo2019design} combined convolutional neural networks with reinforcement learning to construct richer board representations, while Zhang et al. \citep{zhang2021mastering} decomposed the model into a rule network and a skill network to improve learning efficiency and alleviate dependence on computing resources. These methods have shown promise, but they also reveal two common limitations: first, strong performance usually depends on sufficient training data and computational resources; second, representation power is often improved at the expense of interpretability.

Monte Carlo Tree Search has become a standard framework for sequential decision-making in complex games due to its flexibility and its ability to balance exploration and exploitation \citep{kim2024surrogate,swiechowski2023monte,moon2022diversifying,crippa2022analysis}. Compared with exhaustive deterministic search, MCTS can allocate computational effort more adaptively, making it particularly attractive in large search spaces.

However, under limited computational budgets, MCTS still faces practical difficulties. Shallow search may lead to unstable value estimates, while directly increasing the number of explored nodes or search depth substantially raises runtime cost. In games such as Amazons, where a single turn consists of both movement and barrier placement, the search tree expands especially quickly. Therefore, budget-aware improvements to search strategies are critical for maintaining acceptable performance in lightweight deployment scenarios.

Existing studies have mainly focused on improving local node selection, rollout quality, or search efficiency. By contrast, relatively less attention has been paid to exploiting the structural information of the search graph itself for candidate refinement under restricted computational resources. This gap motivates the present work.

Graph-based learning provides a natural way to model relational dependencies among candidate decisions. Unlike vector-based representations, graph neural networks can explicitly capture neighborhood interactions, topological dependencies, and information propagation across connected states. In complex system modeling, for instance, LLMs have been utilized to assist in generating initial weak signals that are subsequently transformed into fine-grained graph networks for precise structural reasoning \citep{llm_graph_2025}. In particular, Graph Attention Networks can adaptively assign different weights to neighboring nodes, making them suitable for decision problems in which not all candidate relationships contribute equally.

Meanwhile, autoencoders offer a lightweight mechanism for feature compression and reconstruction. In resource-constrained settings, compact latent representations are attractive because they may improve feature quality without introducing excessive computational overhead. This property is valuable for game-state evaluation, where representation refinement is needed but inference latency must remain low.

Related ideas have appeared in broader game and evaluation research. For example, Kagkas et al. \citep{kagkas2023chess} proposed an estimator based on a radial basis neural network for robust high-dimensional game-state representation, and Hsueh et al. \citep{hsueh2024improvement} designed a new situation-based smoothing function to calculate move scores in Go. Nevertheless, existing approaches usually evaluate candidate moves independently or rely on handcrafted heuristics, without explicitly transforming the MCTS candidate structure into a graph for structural filtering. Meanwhile, recent advances in engineering informatics have successfully integrated autoencoders and graph attention to process structural synthetic data under severely constrained environments \citep{vae_gat_2024}. This motivates our integration of graph attention and autoencoder-based refinement within the Amazons decision framework.

Large language models have recently been used as scalable generators of synthetic supervision, especially in scenarios where expert annotations are scarce or expensive. This idea is particularly attractive in domains lacking sufficient task-specific training data. At the same time, the notion of weak-to-strong generalization has drawn increasing attention, as it suggests that a weaker or noisier supervisor can still provide useful signals from which a stronger specialized model may emerge \citep{burns2023weak}.

For Amazons, the scarcity of expert game logs makes this perspective especially relevant. Instead of relying on large volumes of historical records, one may attempt to extract useful strategic priors from a general-purpose LLM. However, LLM-generated supervision is inherently imperfect. In structured decision tasks, such supervision may contain hallucinations, inconsistent scores, or noisy judgments, making it unsuitable for direct use without additional filtering or structural correction. This aligns with recent benchmark evaluations, such as AECBench, which highlight the severe limitations and hallucination risks of LLMs when dealing with complex, rule-based spatial and structural reasoning \citep{chen2025aecbench}. Recent comparative analyses also confirm that hybridizing LLMs with graph structures is essential for effectively denoising these imperfect generative outputs \citep{hybrid_llm_graph_2025}.

Therefore, an important open problem is how to exploit the informative part of synthetic supervision while suppressing its errors. Our work addresses this issue by combining stochastic graph selection and graph-attention-based structural filtering, thereby enabling the student model to learn robust decision strategies from imperfect weak supervision.

\section{Preliminaries}
\label{preliminaries}

In this section, the Game of the Amazons and the basic backgrounds of autoencoders and graph attention networks are briefly introduced. These preliminaries provide the necessary foundations for the proposed framework presented in Section~\ref{method}.

\subsection{Game of the Amazons}

The Game of the Amazons, invented by Walter Zamkauskas in 1988, is a two-player deterministic board game played on a square grid. In this work, the standard 10$\times$10 board is considered, and each player controls four amazons. A game state can be represented as
\begin{equation}
S = \langle C_w, C_b, B \rangle,
\end{equation}
where \(C_w = [C_{w_1}, C_{w_2}, C_{w_3}, C_{w_4}]\) denotes the positions of the four white amazons, \(C_b = [C_{b_1}, C_{b_2}, C_{b_3}, C_{b_4}]\) denotes the positions of the four black amazons, and \(B\) denotes the set of barrier positions on the board.

\begin{figure}[H]
\centering
\includegraphics[width=0.4\linewidth]{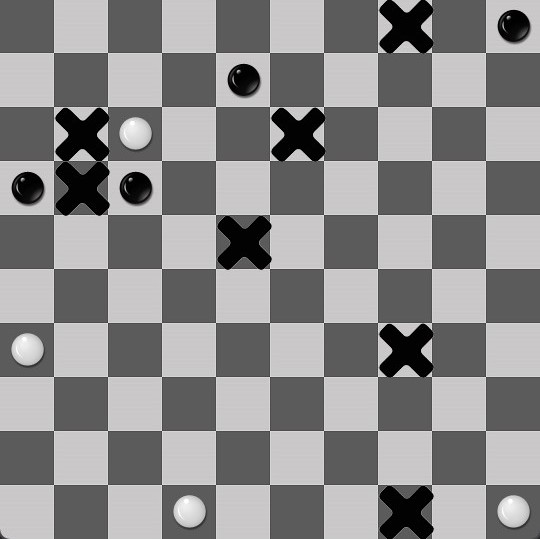}
\caption{An example board configuration in the Game of the Amazons.}
\label{board}
\end{figure}

An example of an intermediate board configuration is shown in Fig.~\ref{board}, where \textbf{X} represents a barrier.

Let \(S_i\) denote the board state at the \(i\)-th turn, and let the sequence of states from the beginning of the game to the current turn be written as
\begin{equation}
S^i = [S_0, S_1, \dots, S_i].
\end{equation}
At each turn, a player first moves one of the four amazons along a straight line, following the same movement rule as the queen in chess, and then places a barrier from the new position according to the same rule. Once placed, a barrier permanently blocks the corresponding grid. The game terminates when a player has no legal move available, in which case that player loses.

Due to the two-stage action rule and the large number of legal actions in each turn, Amazons naturally induces a high-dimensional search space and serves as a challenging benchmark for intelligent decision-making algorithms.

\subsection{Monte Carlo Tree Search}

Monte Carlo Tree Search (MCTS) is a widely used heuristic search framework for sequential decision-making problems with large action spaces. Compared with exhaustive tree search, MCTS incrementally constructs a partial search tree by repeatedly sampling promising actions, which makes it particularly suitable for complex board games such as the Game of the Amazons.

\begin{figure}[H]
\includegraphics[width=0.6\linewidth]{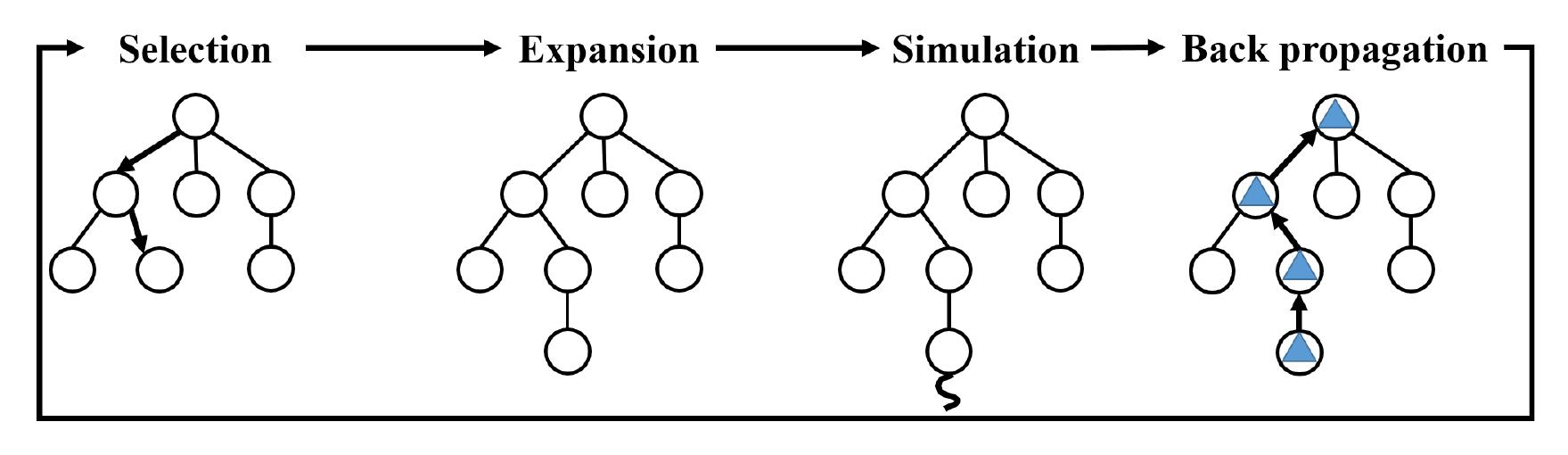}
\centering
\caption{The structure of MCTS.}
\label{UCT}
\end{figure}

Starting from the current root state, MCTS proceeds through four standard steps: selection, expansion, simulation, and backpropagation. The overall process is illustrated in Fig.~\ref{UCT}.

\textbf{Selection.}
Starting from the root node, the algorithm recursively selects the most promising child node according to a predefined policy until a node that is not fully expanded is reached.

\textbf{Expansion.}
If the selected node is non-terminal and still has unexplored actions, one or more child nodes are added to the search tree.

\textbf{Simulation.}
Starting from the expanded node, a rollout is performed until a terminal state or a predefined stopping condition is reached. The simulation outcome provides an estimated reward for the selected trajectory.

\textbf{Backpropagation.}
The simulation result is propagated backward along the visited path so that the visit counts and cumulative rewards of the involved nodes can be updated.

A common tree policy used in MCTS is the Upper Confidence Bound applied to Trees (UCT), which balances exploitation and exploration through
\begin{equation}
\mathrm{UCT}(v_j)
=
\bar{X}_j
+
C\sqrt{\frac{\ln N}{n_j}},
\end{equation}
where \(\bar{X}_j\) is the average reward of child node \(v_j\), \(n_j\) is the visit count of node \(v_j\), \(N\) is the visit count of its parent node, and \(C\) is the exploration constant. The first term favors nodes with high empirical rewards, whereas the second encourages exploration of less-visited nodes. By iteratively repeating these four steps, MCTS gradually allocates more computational effort to promising regions of the search space.

\subsection{Autoencoder}

\begin{figure}[H]
\centering
\includegraphics[width=0.4\linewidth]{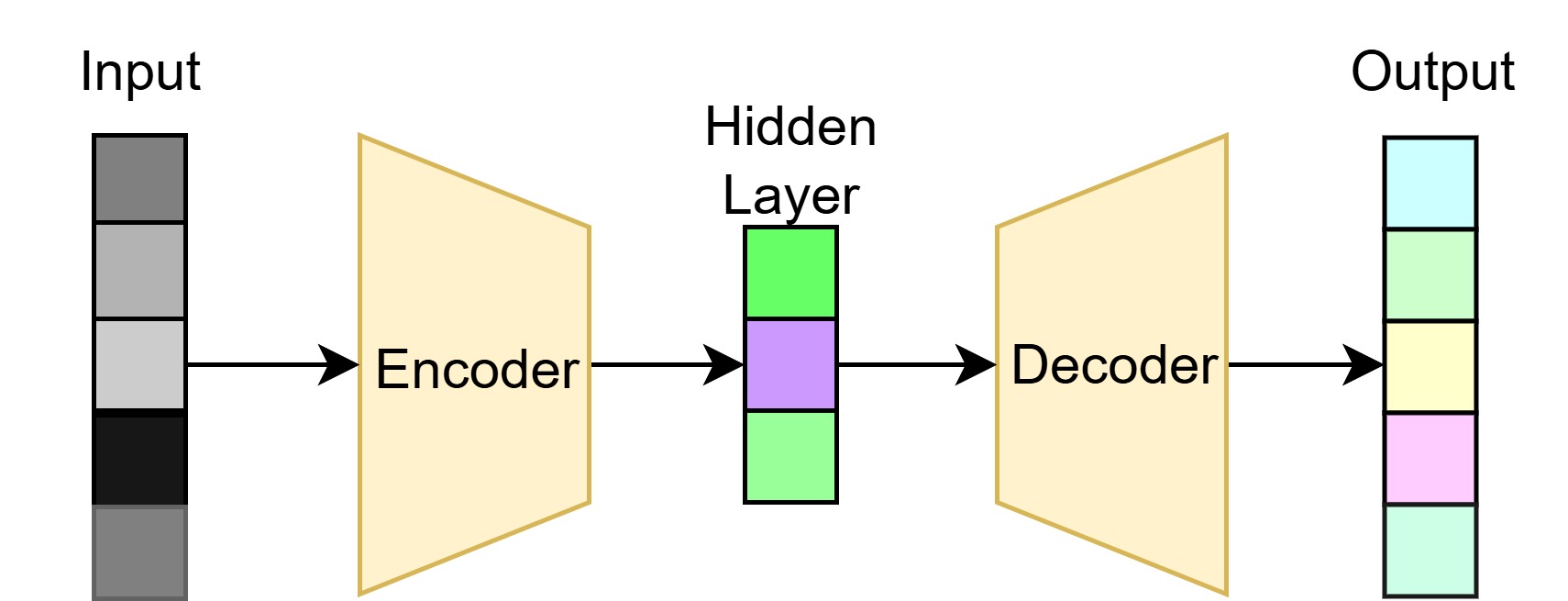}
\caption{The general structure of an autoencoder.}
\label{autoencoder}
\end{figure}

An autoencoder (AE) \citep{2022arXiv220103898M} is an unsupervised neural architecture consisting of an encoder and a decoder. The encoder maps an input vector \(\mathbf{x}\) into a lower-dimensional latent representation \(\mathbf{z}\), and the decoder reconstructs it as \(\hat{\mathbf{x}}\). The objective is to make the reconstructed output as close as possible to the original input. Its general structure is illustrated in Fig.~\ref{autoencoder}.

Formally, the encoder and decoder can be denoted by two mappings
\begin{equation}
A:\mathbb{R}^{n}\rightarrow\mathbb{R}^{p}, \qquad
B:\mathbb{R}^{p}\rightarrow\mathbb{R}^{n},
\end{equation}
where \(n\) is the input dimension and \(p\) is the latent dimension. The autoencoder is trained by minimizing the reconstruction loss
\begin{equation}
\arg\min_{A,B}\ \mathbb{E}\!\left[\Delta\!\left(\mathbf{x},\, B(A(\mathbf{x}))\right)\right],
\end{equation}
where \(\mathbb{E}[\cdot]\) denotes the expectation over the input distribution and \(\Delta(\cdot,\cdot)\) is the reconstruction error, commonly implemented as the \(l_2\)-norm or mean squared error \citep{bank2023autoencoders}. Autoencoders are widely used for nonlinear feature compression, denoising, and latent representation learning. 

\subsection{Graph Attention Networks}

Graph Attention Networks (GATs) \citep{velickovic2018graphattentionnetworks} extend graph neural networks by introducing an attention mechanism over neighboring nodes. Compared with ordinary vector-based models, GATs are able to exploit relational information encoded in graph structures and adaptively assign different importance weights to different neighbors.

Let \(\vec{h}_i\) denote the feature vector of node \(i\), and let \(\mathbf{W}\) be a learnable linear transformation matrix. For node \(i\) and one of its neighbors \(j\), an unnormalized attention coefficient is computed as
\begin{equation}
e_{ij} = \alpha(\mathbf{W}\vec{h}_i,\mathbf{W}\vec{h}_j),
\end{equation}
where \(\alpha(\cdot,\cdot)\) is a shared attention function, typically implemented by a single-layer feedforward network followed by a nonlinear activation.

To ensure comparability among neighbors, the coefficients are normalized by the softmax operation:
\begin{equation}
\alpha_{ij}
=
\mathrm{softmax}_{j}(e_{ij})
=
\frac{\exp(e_{ij})}{\sum\limits_{k\in\mathcal{N}_i}\exp(e_{ik})},
\end{equation}
where \(\mathcal{N}_i\) denotes the set of neighboring nodes of node \(i\).

The updated representation of node \(i\) is then obtained by aggregating the transformed neighbor features:
\begin{equation}
\vec{h}_i^{\,\prime}
=
\sigma\!\left(
\sum\limits_{j\in\mathcal{N}_i}
\alpha_{ij}\mathbf{W}\vec{h}_j
\right),
\end{equation}
where \(\sigma(\cdot)\) denotes an activation function. Through this attention-based aggregation, GATs can capture both local structural dependencies and heterogeneous contributions from neighboring nodes.

Since the proposed framework constructs graph-structured candidate relationships from the search process, GATs provide a natural tool for exploiting such relational information. 
\section{Proposed Framework}
\label{method}

\begin{figure}[H]
\includegraphics[width=0.75\linewidth]{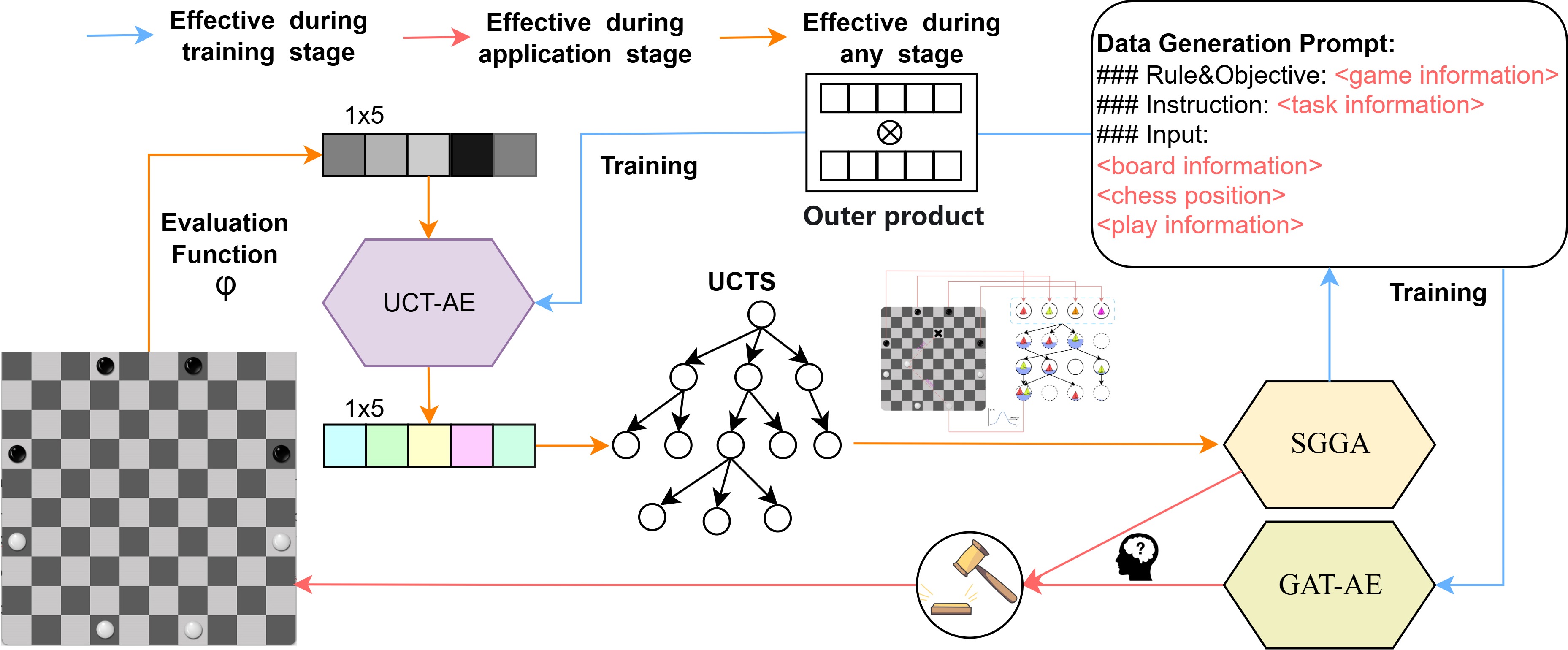}
\centering
\caption{Overall framework of the proposed method.}
\label{framework}
\end{figure}

The proposed framework consists of two stages: a training stage and an application stage. In the training stage, synthetic supervision generated by a large language model is used to train two lightweight evaluation modules, namely UCT-AE and GAT-AE, with the assistance of the Stochastic Graph Genetic Algorithm (SGGA). In the application stage, these trained modules are integrated into the search-and-decision pipeline for movement and placement selection.

From a functional perspective, the two modules are designed to be complementary. UCT-AE improves node evaluation during tree search from the perspective of lightweight feature refinement, whereas GAT-AE further enhances candidate discrimination by exploiting the graph structure induced by MCTS. Meanwhile, SGGA introduces stochastic exploration over graph-structured candidates, thereby improving the diversity and robustness of candidate selection. The overall workflow is illustrated in Fig.~\ref{framework}, and the pseudo-code of the complete framework is provided in Algorithm~\ref{alg:ai_move_placement}.

\begin{breakablealgorithm}
\begin{flushleft}
\caption{AI Move and Placement Strategy (MCTS + SGGA + GAT-AE)}
\end{flushleft}
\label{alg:ai_move_placement}
\begin{algorithmic}[1]
\Require Initial board state $\mathcal{B}_0$; time limit $T_{\max}$; SGGA crossover rate $\sigma$
\Ensure Selected move $m^*$

\State Initialize MCTS tree $\mathcal{T}\leftarrow \mathrm{root}(\mathcal{B}_0)$
\State Load Autoencoder on CPU; load GAT on GPU
\State Set $t\leftarrow 0$; start timer; set mode $m\in\{0,1\}$ (BLACK/WHITE)

\While{$t < T_{\max}$}
    \State $\mathcal{C}\leftarrow \mathrm{UCT\text{-}AE\_Select}(\mathcal{T})$
    \State Update board and tree: 
    \Statex \hspace{\algorithmicindent} $\mathcal{B}\leftarrow \mathcal{B}\cup n_{\text{new}}$, \ $\mathcal{T}\leftarrow \mathrm{insert}(\mathcal{T},n_{\text{new}})$
    \State $t\leftarrow \mathrm{elapsed}()$
\EndWhile

\State Perform SGGA
\State $n_{\text{new}}\leftarrow \mathrm{SGGA}(\sigma,\mathcal{T},\mathcal{COUNTER})$

\Repeat
    \State $n_{\text{new}}\leftarrow \mathrm{SGGA}(\sigma,n_{\text{new}},\mathcal{COUNTER})$
\Until{$\mathrm{NodeCount}(n_{\text{new}})\ge 2^{(\mathrm{Height}_{\max}-\mathrm{NodeHeight}(n_{\text{new}})+1)}$}
\Statex \hspace{\algorithmicindent} \textbf{or} $\mathcal{COUNTER}\ge 50000$

\State Choose node $n^* = n_{\text{new}}$ or $\textbf{None}$
\State Construct subgraph $(X,A)\leftarrow \mathrm{ExtractGraph}(n^*)$
\State $y\leftarrow \mathrm{GAT\text{-}AE}(X,A)$
\State $m^*\leftarrow \arg\max_i y_i$
\Return $\mathrm{Decision\_Strategy}(m^*,n^*)$

\end{algorithmic}
\end{breakablealgorithm}

\subsection{Evaluation Features}
\label{evalfeatures}

To characterize board states with lightweight yet interpretable descriptors, five evaluation features are constructed in this work. Although manually designed features cannot fully represent the complete state space, they provide an informative basis for subsequent refinement by the proposed models. Four features are inspired by value- and knowledge-based domains, while one additional feature is adapted from prior research. These features are defined as follows:

\begin{enumerate}
    \item \textbf{Adjacency-Territory}:  
    This feature measures the ratio of squares that can be reached more quickly by the player than by the opponent when movement is restricted to adjacent-square steps.

    \item \textbf{Line-Territory}:  
    This feature measures the ratio of squares that can be reached more quickly by the player than by the opponent when movement is based on straight-line reachability.

    \item \textbf{One-Mobility}:  
    This feature quantifies the number of adjacent squares that the moved piece can immediately reach. The value is normalized by dividing it by 8.

    \item \textbf{Line-Mobility}:  
    This feature calculates the number of reachable squares for all four pieces after a barrier is placed. The value is normalized by dividing it by 4.

    \item \textbf{Position}:  
    The positional feature is defined as
    \begin{flalign}
    &&
        p = 
        \begin{cases} 
        \sum\limits_{A} \left( 2^{D_1^1(A) - D_1^2(A)} \right), & \text{if } \text{turn} \leq 30, \\
        \sum\limits_{A} \left( 2^{D_2^1(A) - D_2^2(A)} \right), & \text{if } \text{turn} > 30,
        \end{cases}
    &&
    \end{flalign}
    where the formulation is adapted from \citep{lieberum2005evaluation}. Here, \( D_1(A) \) denotes the minimum number of steps required for a square \( A \) to be reached under straight-line movement, and \( D_2(A) \) denotes the corresponding value under adjacent-square movement. The superscripts \(1\) and \(2\) refer to the two competing sides.
\end{enumerate}

Collectively, these five features describe territorial control, local mobility, global mobility, and positional advantage, thereby providing a compact representation of board states for subsequent evaluation refinement.

\subsection{Training Stage}

The training stage aims to learn lightweight evaluation and candidate-selection modules from LLM-generated synthetic data. Starting from the manually constructed evaluation features, the proposed framework refines state representation through autoencoder-based remapping and further exploits the search-tree structure through graph-based candidate modeling. The full training pipeline consists of four major components: MCTS with an update mechanism, UCT-AE, GAT-AE, and SGGA, followed by LLM-based data generation.

\subsubsection{MCTS with Update Mechanism}

MCTS serves as the search backbone of the proposed framework. Its standard procedure has been introduced in Section~\ref{preliminaries}. In this work, an additional update mechanism is incorporated to improve the comparability of nodes at different depths and to reduce the effect of accumulated estimation noise in deep branches.

Hayes et al. \citep{hayes2022montecarlotreesearch} proposed Distributional MCTS (DMCTS), which maintains a posterior distribution over value estimates at each node and employs Thompson sampling for decision-making. Inspired by this distributional perspective, and considering that deeper nodes may accumulate larger variance due to compounding estimation errors, we instead introduce a global depth normalization mechanism. This mechanism rescales node values according to depth, maps the objective values into the interval \([0,1]\), and makes nodes from different layers more directly comparable.

Specifically, a two-stage recursive value-propagation scheme is adopted. Let each node \(n\) carry an objective value \(\mathit{obj}(n)\), let \(\mathit{height}(n)\) denote its depth, and let \(\mathit{H}_{\max}\) denote the overall maximum depth of the tree.

\paragraph*{1. Depth-Dependent Accumulation (Pass I)}
Starting from the leaves and proceeding upward, each internal node computes the arithmetic mean of its children:
\begin{flalign}
    &&
    \bar{v}(n) = \frac{1}{|\mathit{Ch}(n)|} \sum_{c \in \mathit{Ch}(n)} \mathit{obj}(c).
    &&
\end{flalign}

Then, the node value is updated according to the parity of its depth:
\begin{flalign}
    &&
    \mathit{obj}(n) \leftarrow
    \begin{cases}
        \mathit{obj}(n) + \bar{v}(n), & \text{if } \mathit{height}(n) \text{ is even}, \\
        \mathit{obj}(n) + 2^{-\bar{v}(n)}, & \text{if } \mathit{height}(n) \text{ is odd}.
    \end{cases}
    &&
\end{flalign}

\paragraph*{2. Global Depth Normalization (Pass II)}
After the first pass, all node values are rescaled by
\begin{flalign}
    &&
    \mathit{obj}(n) \leftarrow \frac{\mathit{obj}(n)}{\mathit{H}_{\max} + 1 - \mathit{height}(n)}.
    &&
\end{flalign}

This normalization introduces a depth-based penalty to suppress the propagation of unreliable deep-node estimates and to prevent the search process from overcommitting to distant branches with high uncertainty.

\subsubsection{UCT-AE}

The UCT-AE module is designed to refine the handcrafted evaluation features during search while preserving low computational cost. In this work, the input is a \(1\times5\) tensor \(\mathbf{v}\), representing the five evaluation features at the current turn. The tensor is first mapped from the 5-dimensional space to a 3-dimensional latent space through a linear layer, followed by a ReLU activation. It is then reconstructed back to a 5-dimensional representation through another linear layer followed by a Tanh activation. The resulting output is denoted by \(\mathbf{v}'=\mathbf{AE}(\mathbf{v})\).

Because Amazons consists of two sequential actions, namely movement and barrier placement, two distinct autoencoders are employed:
\[
\mathbf{AE}=[\mathrm{AE}_1,\mathrm{AE}_2],
\]
where \(\mathrm{AE}_1\) is used for movement evaluation and \(\mathrm{AE}_2\) is used for placement evaluation.

Based on the refined representation, the modified upper confidence bound of the \(j\)-th node is defined as
\begin{flalign} 
&&
\mathbf{UCB}_j = \mathbf{AE}_1(\mathbf{v}_j)\mathbf{W}_1  + \sqrt{\frac{2 \ln n}{n_j + 1}},
&&
\end{flalign}
where \(n\) is the total number of simulations and \(n_j\) is the number of times the \(j\)-th node has been selected.

The overall evaluation is then computed as
\begin{flalign}
&&
v_j = \alpha \cdot \mathbf{AE}_1(\mathbf{v}_j)\mathbf{W}_1  + (1 - \alpha) \cdot \mathbf{AE}_2(\mathbf{v}_j)\mathbf{W}_2  + \sqrt{\frac{2 \ln n}{n_j + 1}},
&& \label{AE-1}
\end{flalign}
where \(\alpha\) is an adjustable parameter. In this way, lightweight representation refinement is embedded directly into the UCT-based search process.

\subsubsection{GAT-AE}
\label{gat_ae}

The GAT-AE module is introduced to exploit the structural information induced by the MCTS tree. Since standard GAT requires both a node feature matrix and an adjacency matrix, the candidate tree is first transformed into a graph representation. To reduce redundancy, the four root nodes are temporarily aggregated into a super-node.

Let the node feature input be denoted by \(\mathbf{X}\). The input is first processed by the feature-refinement mechanism in Eq.~\eqref{AE-1}, after which graph attention layers are applied. The final output is passed through a shifted activation:
\begin{flalign}
    &&
    \mathbf{O}=f(v_j)=\frac{\tanh(v_j)+1}{2}.
    &&
\end{flalign}

This transformation maps raw scores into the interval \([0,1]\) and sharpens the distinction between favorable and unfavorable candidates. Such a non-linear normalization is helpful for the subsequent SGGA procedure, as it enhances selection pressure without requiring gradient-based optimization during graph traversal.

\begin{figure}[H]
\includegraphics[width=0.6\linewidth]{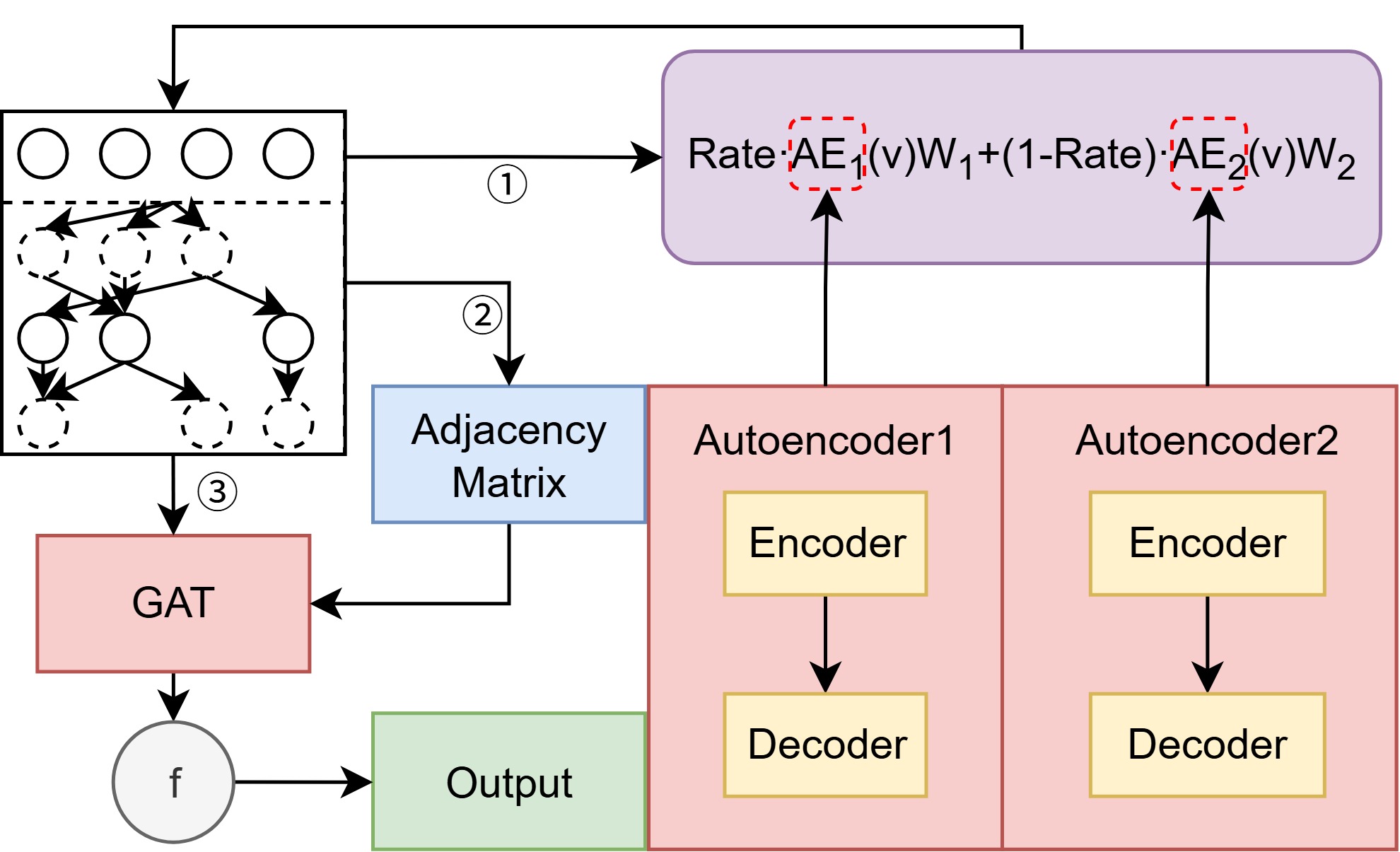}
\centering
\caption{The structure of GAT-AE.}
\label{GAT}
\end{figure}

\subsubsection{Stochastic Graph Genetic Algorithm (SGGA)}

The SGGA module is introduced to perform stochastic candidate selection on the graph transformed from the MCTS tree. First, the four head nodes corresponding to the four amazons are connected with undirected edges, so that the search tree is converted into a graph. Then, a genetic procedure is employed to explore this graph from a probabilistic perspective.

The SGGA consists of three operations:

\begin{enumerate}
\item \textbf{Selection.}  
Two candidate nodes are sampled from a genetic repository according to a softmax distribution. The repository stores the history of generated nodes and is initialized with two candidates.

\item \textbf{Mutation.}  
A biased random walk \citep{10.1145/2939672.2939754} is used to generate the next candidate. Given mutation step size \(\sigma=0.8\), the next node is chosen from the children with probability \(0.8\) and from the parent with probability \(0.2\).

\item \textbf{Crossover.}  
When two candidates arrive at the same position, crossover is triggered and one additional candidate is added to the repository.
\end{enumerate}

The algorithm terminates when the count of any node satisfies
\begin{flalign}
&&
    \textbf{NodeCount} 
    \;\ge\ 
    2^{\bigl(\textbf{Heightmax} - \textbf{NodeHeight} + 1\bigr)}.
&&
\end{flalign}
An upper bound of 50000 generations is imposed to avoid dead-end loops. After a target node is identified, its trajectory is traced back to recover the corresponding movement and placement decision.

\begin{figure}[H]
\includegraphics[width=0.7\linewidth]{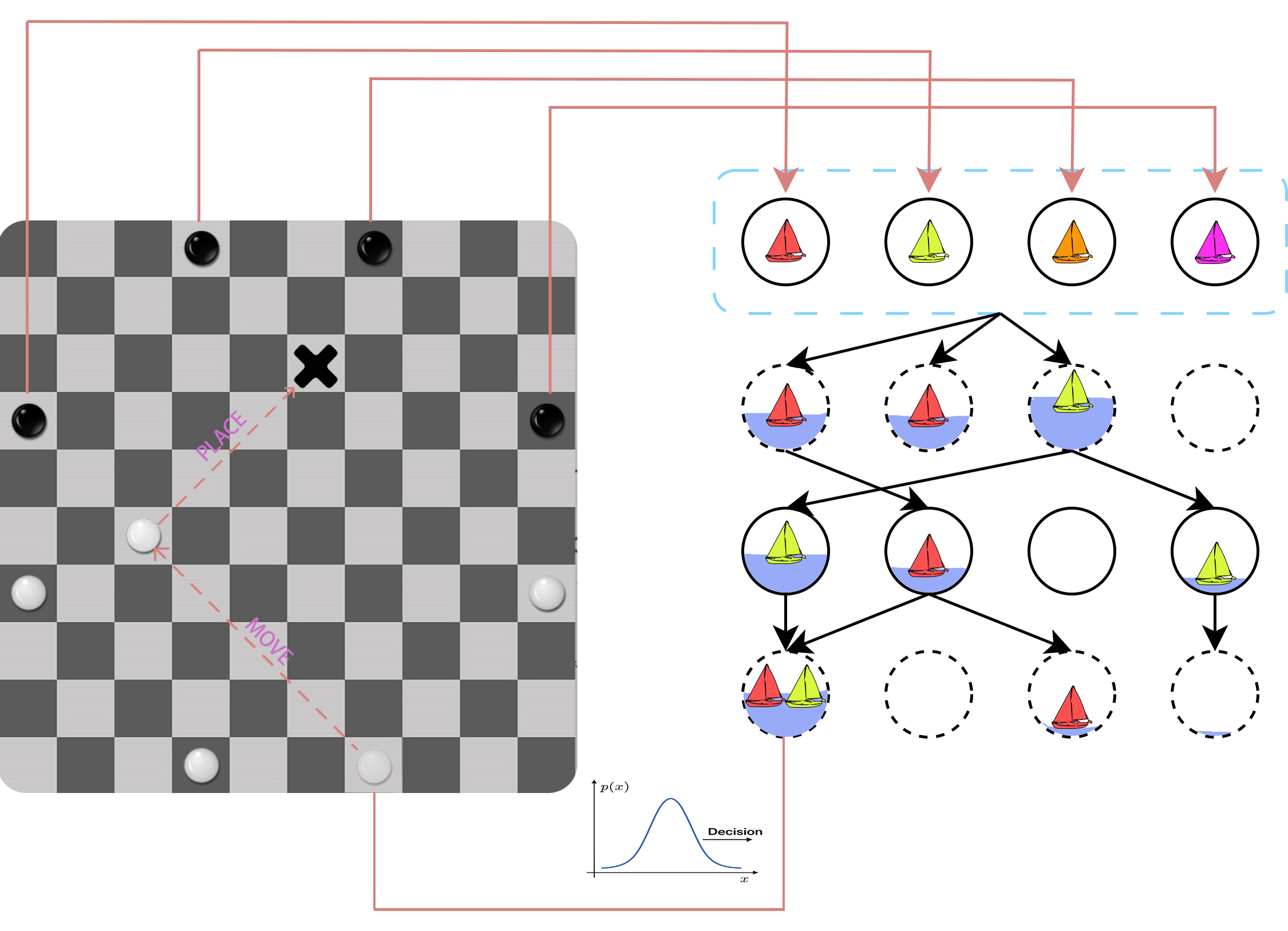}
\centering
\caption{The structure of SGGA.}
\label{SGGA}
\end{figure}

\subsubsection{LLM-based Training Data Generation}

To avoid dependence on scarce expert game logs, synthetic supervision is generated by a general-purpose large language model, specifically GPT-4o-mini. Given the board configuration and candidate moves, the LLM provides score annotations for candidate actions.

However, LLM-generated scores may be noisy, inconsistent, or difficult to represent precisely as stable floating-point values. To mitigate this issue, the proposed framework does not directly rely on these scores as exact labels. Instead, SGGA is used to decompose the generated information into graph-based probabilistic selection signals, which are then used to train the downstream deep models more robustly.

In this way, LLM-generated supervision serves as a weak supervisory source, while the proposed hybrid framework learns task-specific representations and structural decision patterns from it.

\subsection{Application Stage}

In the application stage, the trained UCT-AE and GAT-AE modules are integrated into the online decision-making process for the Game of the Amazons. UCT-AE is first used within MCTS to maintain the balance between exploitation and exploration during candidate expansion. Then, SGGA and GAT-AE are jointly applied to the graph-structured candidate space in order to identify the final movement and placement decision from both stochastic and structural perspectives.

At the current stage, the final decision is made according to the overall objective value of the candidate node, with random tie-breaking when necessary. A more principled decision strategy will be investigated in future work.

\section{Experiment and Analysis}
\label{experiment}

This section evaluates the proposed framework from four perspectives. First, the experimental setup is introduced, including the hardware environment, evaluation protocol, and baseline settings. Second, the training dynamics of the movement and placement modules are analyzed through their loss curves. Third, the proposed hybrid model is compared with its teacher model, GPT-4o-mini, under strict search-budget constraints. Fourth, ablation experiments against UCTS-AE, SGGA, and GAT-AE are conducted to examine the contribution and complementarity of each component. Finally, the overall win-rate comparison is summarized.

\subsection{Experimental Setup}

To validate the proposed framework, we implemented a complete Game of the Amazons system and conducted all experiments on a platform equipped with an AMD Radeon(TM) 780M and an NVIDIA GeForce RTX 4060 Laptop GPU. Compared with the hardware typically required by large-model-based systems, this setup is relatively modest, which is consistent with the resource-constrained objective of this work.

All competitive evaluations were conducted in the form of head-to-head games. Unless otherwise specified, each reported win rate was computed over 200 games. For the main comparison against the teacher model GPT-4o-mini, the search budgets were set to \(N=30\) and \(N=50\) in order to examine the minimum search scale at which the proposed framework can become competitive. For the ablation experiments, the search budgets were set to \(N=20\) and \(N=30\), respectively, to test whether the observed performance trends remain stable under different node limits.

The ablation baselines include UCTS-AE, SGGA, and GAT-AE. For the UCTS-AE model, the Adam optimizer with a learning rate of 0.01 and the mean squared error (MSE) loss were adopted. For the GAT-AE model, the RMSprop optimizer with a learning rate of 0.0001 and the Smooth L1 loss were used. In the hybrid framework, the corresponding UCTS-AE and GAT-AE modules share the same internal settings as their standalone counterparts to ensure a fair comparison.

\subsection{Training Dynamics}

\begin{figure}[H]
    \centering
    \subfloat[Training loss of the movement module.]{
        \includegraphics[width=0.47\linewidth]{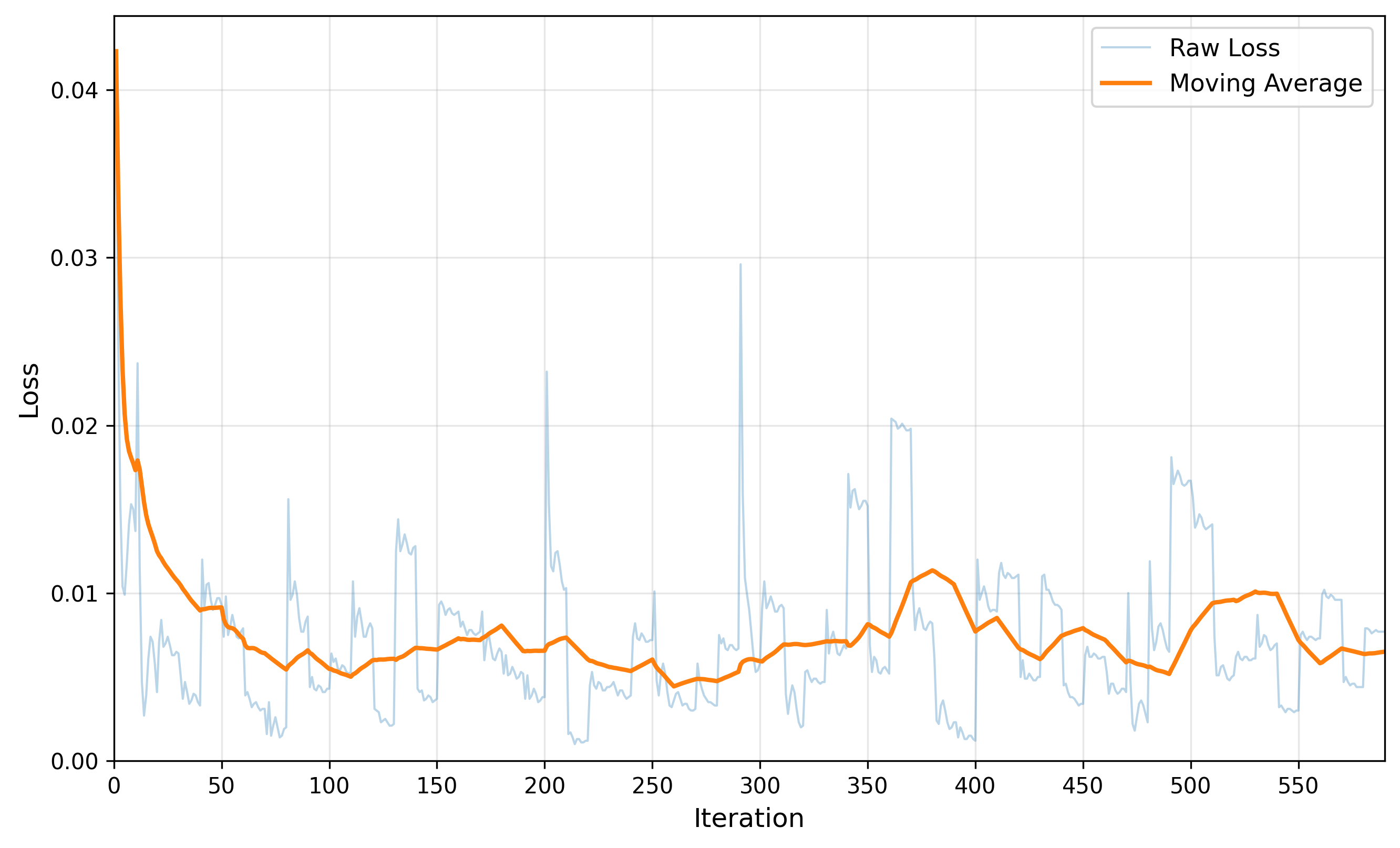}
        \label{loss1}
    }
    \hfill
    \subfloat[Training loss of the placement module.]{
        \includegraphics[width=0.47\linewidth]{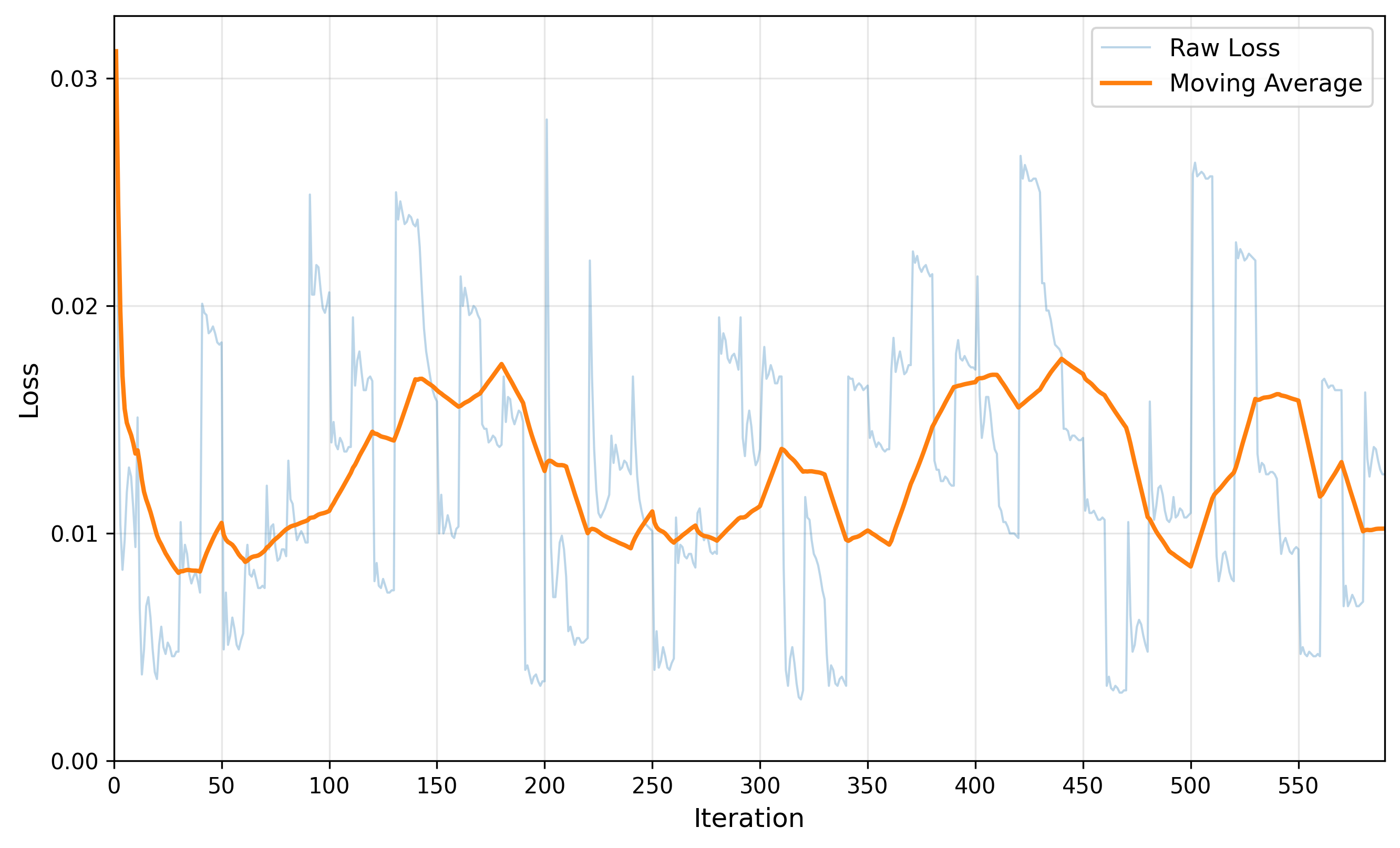}
        \label{loss2}
    }
    \caption{Training loss curves of the movement and placement modules.}
    \label{fig:loss}
\end{figure}

To further examine the convergence behavior and training stability of the two modules, we analyze their loss curves throughout the training process. Figs.~\ref{loss1} and \ref{loss2} present the training losses of the movement and placement modules, respectively.

In both figures, the blue curve denotes the raw MSE loss, while the red curve represents the smoothed trend obtained using a moving average with a window size of 50. Overall, both modules show a clear downward trend, indicating that the proposed framework is able to optimize the two tasks effectively.

For the movement module, the loss decreases rapidly from approximately 0.04 to 0.01 within the first 500 iterations and then gradually enters a relatively stable stage. The placement module exhibits a similar decline in the early stage, with the loss dropping from about 0.03 to 0.01. However, compared with the movement module, the placement module shows more noticeable fluctuations in the later stage, suggesting that its optimization process is relatively less stable.

To further quantify this difference, we compute the variances of the two loss curves after epoch 50. The variance of the movement loss is \(8.0\times10^{-6}\), whereas that of the placement loss is \(2.1\times10^{-5}\). In addition, an F-test yields \(p=0.035\), indicating that the variance difference is statistically significant at the \(\alpha=0.05\) level. These results suggest that the movement module exhibits more stable convergence and lower training variance than the placement module.

One possible reason for this discrepancy is the difference in the data-generation strategies adopted in the two stages. Specifically, movement selection is guided by SGGA, whereas placement decisions are generated using a simpler weighted random strategy. Since the selection mechanism is the major distinction between the two tasks, the lower variance observed in the movement module may indicate that SGGA is associated with more consistent node selection and, consequently, more stable training behavior. Nevertheless, this interpretation remains indirect, and a more controlled ablation study would be needed to establish the effect more conclusively.

\subsection{Main Comparison with the Teacher Model GPT-4o-mini}

To further assess the effectiveness of the proposed hybrid framework under limited computational resources, we compare it with its teacher model, GPT-4o-mini. Since the main focus of this work is resource-constrained decision-making, the purpose of this experiment is not merely to report a higher win rate, but to examine whether the proposed framework can remain competitive with a strong general-purpose model under a small search budget.

\begin{figure}[H]
    \centering
    \includegraphics[width=0.6\linewidth]{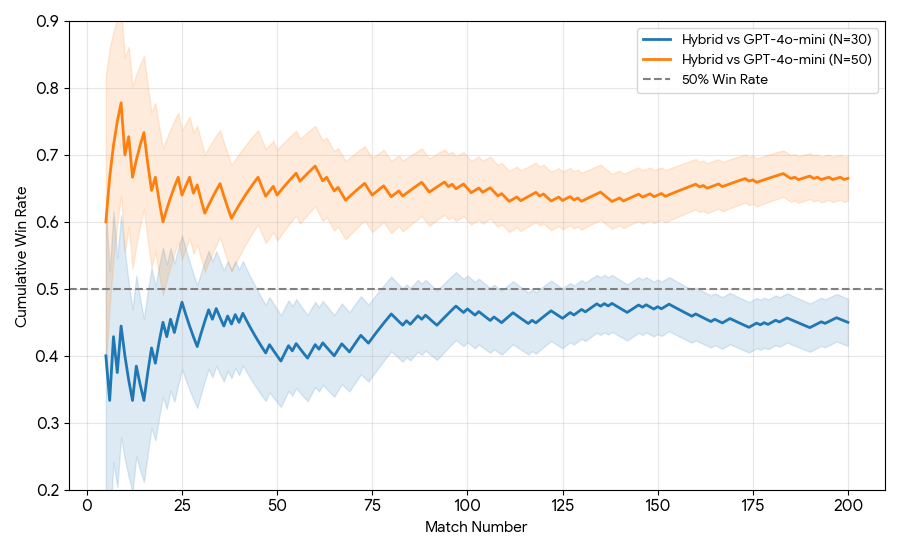}
    \caption{Comparison between the hybrid model and GPT-4o-mini under \(N=30\) and \(N=50\) search budgets.}
    \label{AI}
\end{figure}

As shown in Fig.~\ref{AI}, under the highly restricted search budget of \(N=30\), the proposed model already achieves a competitive win rate of 45.0\%. When the search budget is increased to \(N=50\), the win rate further rises to 66.5\%. These results suggest that the proposed framework can achieve favorable performance against GPT-4o-mini with only a modest increase in search cost, indicating its potential effectiveness in resource-limited settings.

From a broader perspective, these findings suggest that supervision derived from GPT-4o-mini can still support the learning of a competitive task-specific policy for the Game of the Amazons. Although the training signals originate from a general-purpose teacher model, the learned framework appears capable of achieving stronger task-oriented performance when combined with lightweight search.

A qualitative observation is that GPT-4o-mini may require more explicit prompting to consistently track the coordinates of the four amazons and avoid illegal moves. In contrast, the proposed hybrid framework operates directly on graph-structured search states and incorporates stochastic candidate optimization, which may provide a more suitable mechanism for structured board-game reasoning. Therefore, the comparison with GPT-4o-mini provides preliminary empirical evidence that weak-to-strong generalization can be observed in this lightweight game-intelligence setting.
\subsection{Ablation Study}

To examine the necessity and complementarity of the major components in the proposed framework, ablation experiments were conducted against three baselines: UCTS-AE, SGGA, and GAT-AE. Two search budgets, namely \(N=20\) and \(N=30\), were adopted to verify whether the relative advantages of the hybrid framework remain stable under different node limits.

\begin{figure}[htbp]
\centering
\begin{minipage}{0.49\linewidth}
    \centering
    \includegraphics[width=0.9\linewidth]{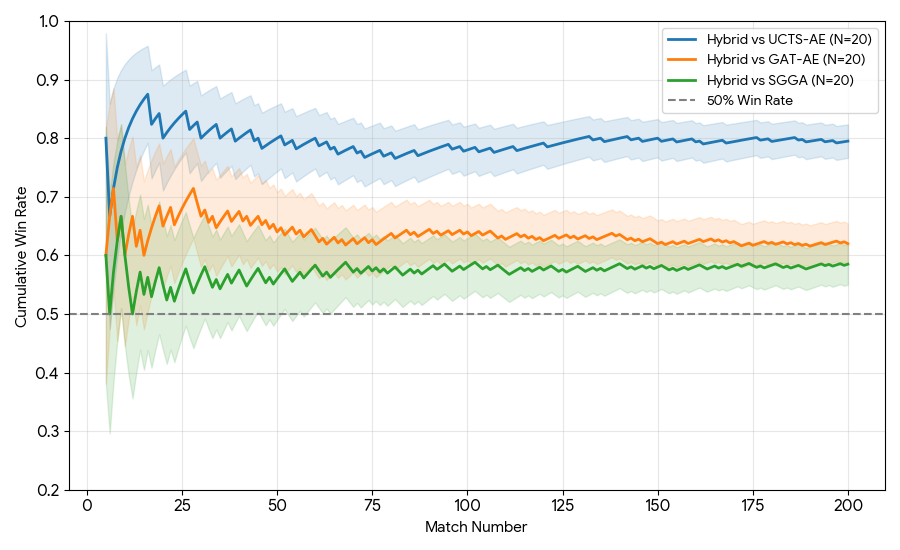}
    \caption{Overall performance of the hybrid model under \(N=20\).}
    \label{fig:N20}
\end{minipage}
\begin{minipage}{0.49\linewidth}
    \centering
    \includegraphics[width=0.9\linewidth]{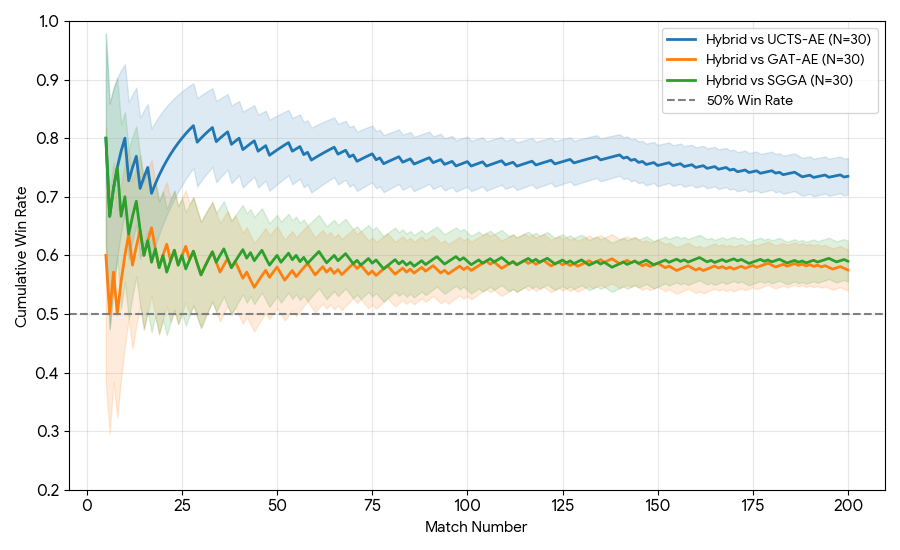}
    \caption{Overall performance of the hybrid model under \(N=30\).}
    \label{fig:N30}
\end{minipage}
\end{figure}

Figs.~\ref{fig:N20} and \ref{fig:N30} summarize the overall win-rate trends of the proposed framework against all three ablation baselines. Under both search budgets, the hybrid model consistently outperforms UCTS-AE, SGGA, and GAT-AE, indicating that the proposed framework does not rely on a single dominant component but instead benefits from the coordinated interaction of lightweight evaluation refinement, stochastic exploration, and graph-structured representation learning. Moreover, the relative ordering of the baselines remains broadly stable across \(N=20\) and \(N=30\), suggesting that the advantage of the hybrid design is not tied to a specific node budget and exhibits reasonable robustness.

\subsubsection{Hybrid Model vs. UCTS-AE}

\begin{figure}[htbp]
    \centering
    \includegraphics[width=0.6\linewidth]{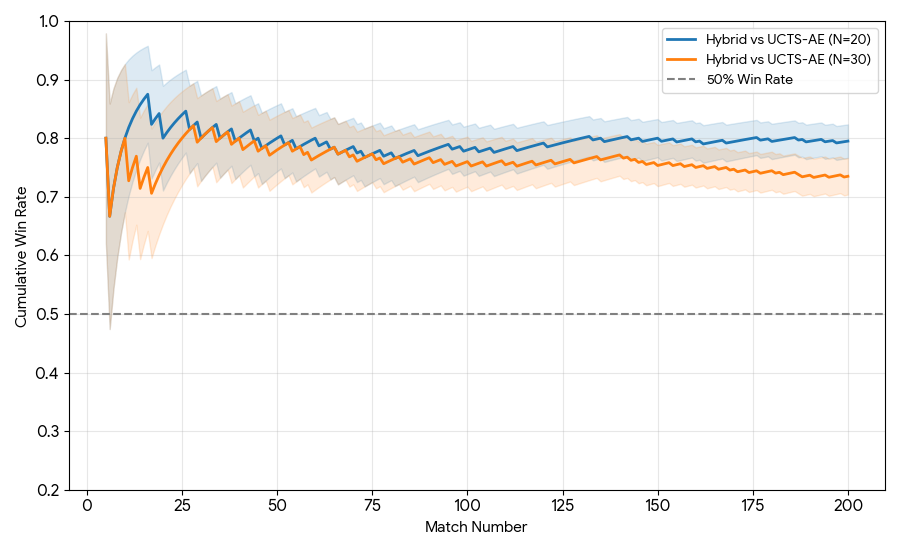}
    \caption{Comparison between the hybrid model and UCTS-AE.}
    \label{fig:HvU}
\end{figure}

UCTS-AE serves as the lightweight nonlinear evaluation module within the overall system. As shown in Fig.~\ref{fig:HvU}, the proposed hybrid model achieves a win rate of 79.5\% (159/200) against UCTS-AE under \(N=20\), and 73.5\% (147/200) under \(N=30\). These results indicate that evaluation refinement alone is insufficient to achieve the full performance of the proposed framework. Even when the same evaluation backbone is retained, the additional incorporation of graph-structured candidate modeling and stochastic exploration leads to a substantial improvement in decision quality.

From a functional perspective, this comparison reveals the limitation of relying only on local score refinement. UCTS-AE can improve node-level evaluation, but it does not explicitly exploit structural dependencies among candidate nodes, nor can it adequately preserve exploration diversity during downstream selection. By contrast, the hybrid framework supplements local evaluation with SGGA-based stochastic traversal and GAT-based structural filtering, thereby enabling more reliable discrimination among competing candidates.

It is also noteworthy that the win rate decreases slightly from \(N=20\) to \(N=30\). This suggests that UCTS-AE may benefit from the larger search budget and partially recover its strength when more candidate nodes become available. Nevertheless, the hybrid model still maintains a clear margin at both budgets, which confirms that the observed gain is structural rather than incidental.

\subsubsection{Hybrid Model vs. SGGA}

\begin{figure}[htbp]
    \centering
    \includegraphics[width=0.6\linewidth]{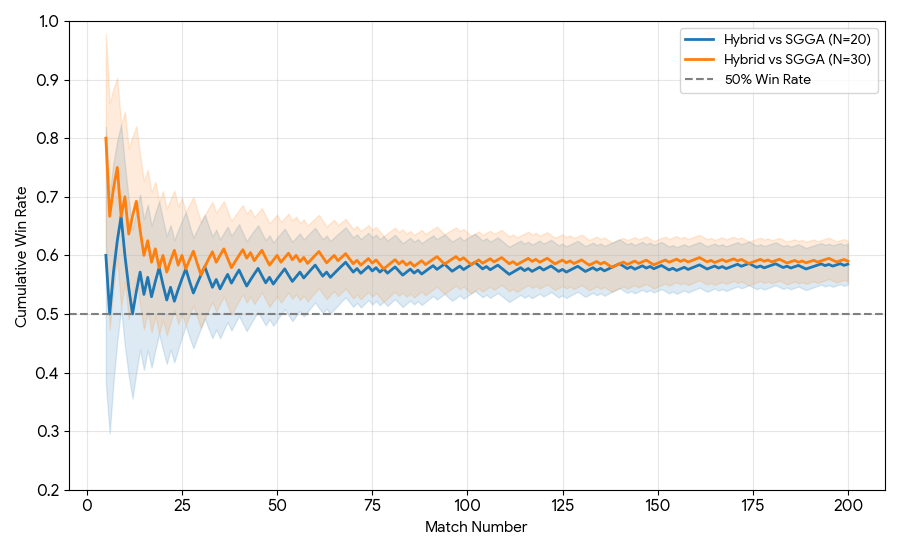}
    \caption{Comparison between the hybrid model and SGGA.}
    \label{fig:HvS}
\end{figure}

SGGA mainly captures local candidate relationships in the Monte Carlo tree from a stochastic perspective. As shown in Fig.~\ref{fig:HvS}, the proposed hybrid model achieves a win rate of 58.5\% (117/200) under \(N=20\) and 59.0\% (118/200) under \(N=30\). Although the improvement margin is smaller than that observed against UCTS-AE, the advantage remains consistent across both search budgets.

This result suggests that stochastic exploration is indeed useful, since SGGA alone already provides a nontrivial search mechanism for candidate discovery. However, its ability is still limited by the lack of an explicit structural evaluation model. In other words, SGGA can increase diversity and reduce the risk of overly deterministic local choices, but it cannot fully assess the broader topological relationships among candidate nodes in the search graph. The hybrid framework overcomes this limitation by coupling SGGA with GAT-AE, so that stochastic exploration is followed by graph-based structural discrimination rather than used in isolation.

Another noteworthy observation is the stability of the win rate across \(N=20\) and \(N=30\). This relatively unchanged gap implies that the advantage of the hybrid model over SGGA is not mainly derived from larger search depth, but from the additional structural information extracted after stochastic exploration. Therefore, this comparison provides direct evidence that stochasticity and structure learning play different but complementary roles in the proposed framework.

\subsubsection{Hybrid Model vs. GAT-AE}

\begin{figure}[htbp]
    \centering
    \includegraphics[width=0.6\linewidth]{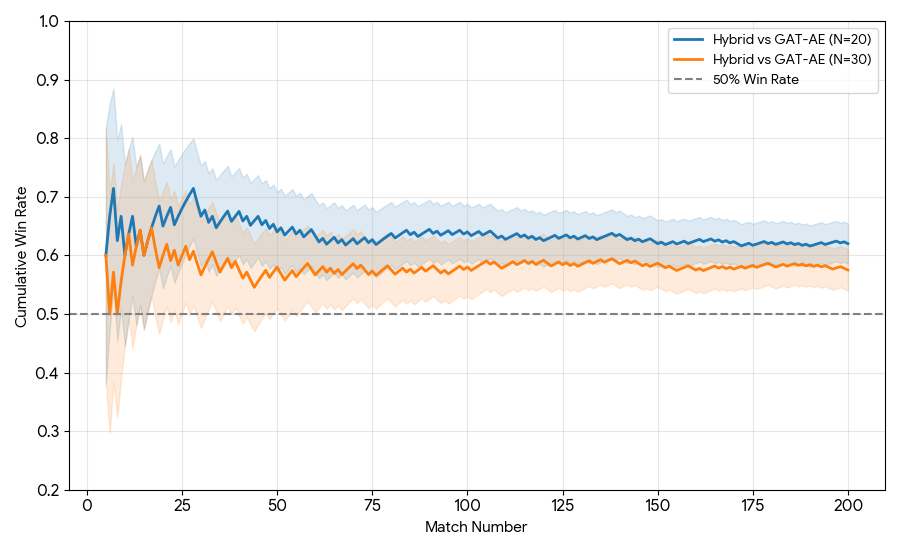}
    \caption{Comparison between the hybrid model and GAT-AE.}
    \label{fig:HvG}
\end{figure}

GAT-AE mainly captures the structural information of the Monte Carlo tree. As shown in Fig.~\ref{fig:HvG}, the proposed hybrid model reaches a win rate of 62.0\% (124/200) under \(N=20\) and 57.5\% (115/200) under \(N=30\). Although the improvement margin is smaller than that observed against UCTS-AE, the hybrid model still consistently outperforms GAT-AE at both search budgets.

This result indicates that structural modeling alone is beneficial but not sufficient. GAT-AE can exploit relational information among candidate nodes and thus improve structural discrimination, but it does not explicitly maintain exploration diversity during the candidate-generation process. As a result, the quality of its final decision still depends on the quality and diversity of the candidate subgraph supplied to it. In the proposed framework, this weakness is alleviated by combining GAT-AE with SGGA and UCTS-AE, so that candidate generation, candidate refinement, and candidate ranking are jointly optimized.

The smaller gap at \(N=30\) may imply that GAT-AE becomes more effective when richer search structures are available, since a larger node budget provides more informative graph topology for attention-based modeling. Even so, the hybrid framework still preserves an advantage, which suggests that structural learning yields the greatest benefit when it is embedded within a broader pipeline rather than used as a standalone component.

Overall, the ablation results support the central design principle of the proposed framework: neither lightweight evaluation refinement, nor stochastic exploration, nor structural modeling alone is sufficient to achieve the best performance. The consistent superiority of the hybrid model across all comparisons indicates that its strength arises from the complementarity of these components, rather than from any single module in isolation.
\subsection{Overall Summary}

The purpose of this subsection is to provide a unified comparison of the proposed framework across all opponents and search budgets, so as to evaluate its overall effectiveness, robustness, and scalability under resource-constrained settings.

\begin{figure}[H]
  \centering
  \includegraphics[width=0.7\textwidth]{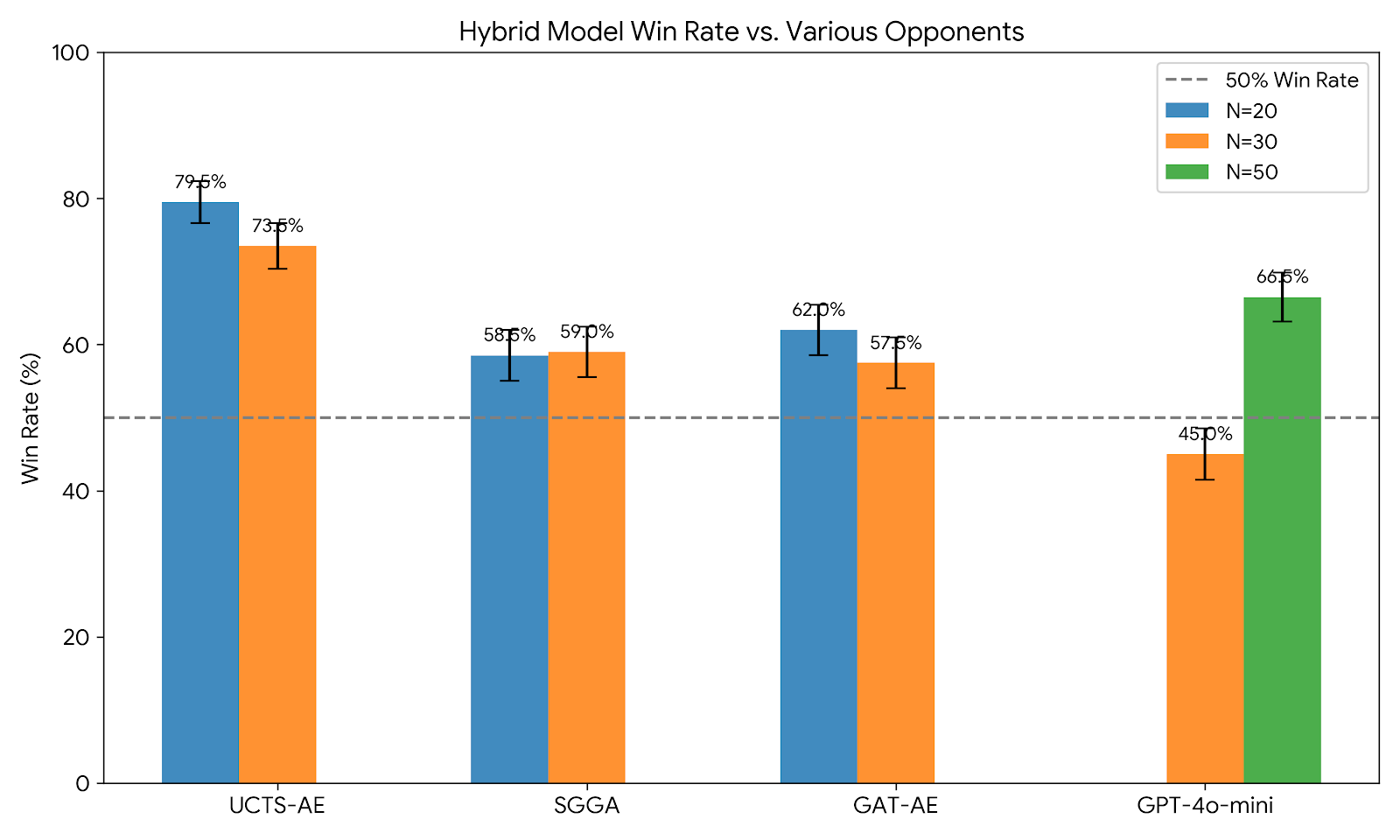}
  \caption{Overall win-rate comparison across different opponents and search budgets.}
  \label{fig:win-rate-comparison}
\end{figure}

Fig.~\ref{fig:win-rate-comparison} summarizes the overall competitive performance of the proposed framework against both the teacher model and the ablation baselines. All results are obtained from 200 games, with confidence intervals computed from the empirical binomial proportions.

Across all ablation settings, the proposed framework consistently outperforms its individual components, achieving win rates between 57.5\% and 79.5\% under \(N=20\) and \(N=30\). Against GPT-4o-mini, the win rate rises from 45.0\% at \(N=30\) to 66.5\% at \(N=50\), indicating that the proposed method can quickly benefit from a moderate increase in search budget when facing a strong general-purpose opponent.

Taken together, these findings verify the effectiveness and robustness of the proposed framework in resource-constrained settings. They also confirm that the performance gain does not arise from a single module alone, but from the complementarity of MCTS-based search, autoencoder-based refinement, SGGA-based stochastic exploration, and GAT-based structural modeling.
\section{Discussion}\label{sec:discussion}

It should be emphasized that established handcrafted engines for Amazons, such as Invader, still represent a higher level of absolute playing strength. However, such systems are typically built upon years of domain-specific heuristic engineering, which makes them highly effective for a particular game but difficult to transfer to other decision-making problems. By contrast, the primary objective of this work is not to surpass all specialized engines in absolute strength, but to investigate whether effective decision policies can be learned from weak and inexpensive supervision in a resource-constrained setting.

From this perspective, the proposed framework provides a different form of value. Instead of relying on expert-designed heuristics or large collections of historical game records, it learns from synthetic signals generated by a general-purpose language model and subsequently improves upon them through structured search and representation learning. The experimental finding that the hybrid framework can outperform GPT-4o-mini in direct competition is consistent with the weak-to-strong generalization phenomenon discussed by Burns et al. \citep{burns2023weak}. More importantly, this result suggests that the proposed framework can extract useful strategic regularities even when the supervisory source is imperfect.

A plausible explanation is that the architectural design itself imposes an inductive bias on the learning process. In particular, SGGA and GAT-AE do not directly imitate the raw outputs of the weak supervisor. Instead, they transform the supervision into graph-structured candidate relationships and topological patterns, thereby reducing reliance on noisy surface-level labels. In this sense, the framework may be viewed not merely as a game-playing system, but as a prototype of a broader learning paradigm for tasks in which expert supervision is unavailable, expensive, or inherently noisy.

A central concern in using LLM-generated data is the presence of hallucinated, inconsistent, or suboptimal supervision. In the context of Amazons, such noise may manifest as illegal moves, unstable evaluations, or strategically weak recommendations. If these raw outputs are used directly as supervision, the downstream model may inherit their errors and exhibit poor generalization.

The experimental results suggest that the graph-based design of the proposed framework plays an important role in mitigating this problem. In particular, the GAT-AE module is constrained to learn from graph-structured candidate relationships rather than from isolated labels alone. As a result, it is encouraged to focus on relatively stable structural patterns, such as connectivity, territorial interaction, and the relational organization of candidate moves, instead of memorizing noisy details in the supervisor output. This structural bias provides a natural form of regularization during training.

Therefore, the proposed framework can be interpreted as performing a form of implicit denoising over weak supervision. Rather than faithfully reproducing every signal provided by the LLM, it selectively preserves the information that is consistent with the search structure and suppresses patterns that are likely to be random or contradictory. This may explain why the student model is able to exceed its weak teacher and why the framework remains robust even when the underlying supervision is imperfect.
\section{Conclusion}

In this paper, a lightweight hybrid framework for resource-constrained intelligent decision-making in the Game of the Amazons was developed by integrating SGGA with GAT-AE. Based on Monte Carlo Tree Search, the proposed method jointly models node values and tree structure, while synthetic supervision generated by a large language model is used to alleviate the scarcity of expert game records. Experimental results showed that the proposed framework consistently outperformed several representative baselines, including GPT-4o-mini, under limited search iterations, demonstrating its effectiveness in low-resource settings.

Future work will focus on two aspects. First, more rigorous criteria for training completion and convergence assessment will be investigated, since loss values alone may not adequately reflect model maturity. Second, a more principled final decision strategy will be developed to replace the current simple stochastic mechanism, with the aim of further improving decision quality and exploring the full potential of the proposed framework.

\bibliographystyle{elsarticle-num-names}
\bibliography{cas-refs}

\newpage
\Huge \textbf{Appendix}

\appendix
\normalsize
\renewcommand{\thesection}{\Alph{section}} 
\renewcommand{\thesubsection}{\Alph{section}\arabic{subsection}} 

\setcounter{section}{0} 

\section{Case Study}\label{Prompt}

\begin{table}[htbp!]
  \centering
  \caption{Case Study of Data Generation}
  \label{tab:three_line_with_prompt}
  \begin{tabular}{@{} p{0.96\textwidth} @{}}
    \toprule
    \multicolumn{1}{@{}l@{}}{\textbf{INPUT}} \\
    \midrule
    $[string]$: Board information
    
    $[chess]$: "white" or "black"\\
    
    $[target]$: 1 if chess== "white" else 2\\
    
    $[step[0]]$: Position of chess to move \\

    $[step[1]]$: Position of chess to reach \\
    
    $[put]$: Position of obstacle to place\\
    \midrule
    \multicolumn{1}{@{}l@{}}{\textbf{PROMPT}} \\
    \midrule
      Amazon is a two-player abstract strategy game that combines elements of strategy and board games. Below are the basic rules of Amazon:
      \\
      \textbf{1.Board and Pieces}

      Board: Amazon is played on a 10×10 grid.

      Pieces: Each player has four “Amazons”, typically distinguished by color (e.g., White vs. Black).
      \\
      \textbf{2.Objective}

      Players aim to occupy as much space as possible by moving their Amazons and firing arrows, while simultaneously blocking the opponent’s mobility.
      \\
      \textbf{3.Rules of Play}

      Initial Setup: Each player’s four Amazons are placed on predetermined squares of the first and last ranks.

      Turn Sequence: Players alternate turns. On your turn, you perform two actions in order:

      \quad Move: Choose one of your Amazons and move it along any straight line—horizontal, vertical, or diagonal—for any number of empty squares, without jumping over other pieces.

      \quad Shoot: After moving, choose a target square along another straight line from that Amazon’s new location; that square becomes permanently blocked and cannot be occupied or traversed.

      Restrictions: You may not move into or shoot at squares that are already occupied or already blocked.
      \\
      \textbf{4.Additional Rule}

      Players must ensure they follow the movement and shooting rules at every step.
      \\
      Please review the above rules. Now you are a professional Amazon player, and the current position is:
      \\
      $[string]$
      \\
      Here, 1 represents White Amazons, 2 represents Black Amazons, and 3 represents blocked squares. You are to evaluate the move just played by player [chess] (ID [target]): they moved the Amazon with index [step[0]] to square [step[1]], then place an obstacle at [put]. Based on both the current and potential future positions, score this turn using two values (each between 0 and 1):
      \\
      $[move\_score\quad place\_score]$
      \\
      A score closer to 1 favors the player; closer to 0 favors the opponent.
      \\
      Please output exactly the above format and no other text. \\
    \midrule
    
    \multicolumn{1}{@{}l@{}}{\textbf{OUTPUT}} \\
    \midrule
    $[move\_score]$: The evaluation score of movement given by LLMs.\\
    
    $[place\_score]$: The evaluation score of placement given by LLMs.\\
    \bottomrule
  \end{tabular}
\end{table}

\end{document}